\documentclass[5p,twocolumn]{elsarticle}

\usepackage{amsmath,amssymb,amsfonts}
\usepackage{algorithmic}
\usepackage{graphicx}
\usepackage{textcomp}
\usepackage{booktabs}
\usepackage{multirow}
\usepackage{hyperref}
\usepackage{xcolor}
\usepackage{subcaption}
\usepackage{float}

\journal{Image and Vision Computing}

\begin{document}

\begin{frontmatter}

\title{A Unified Resolution-Conditioned Framework for Orthogonal Line-Scanning Image Fusion}

\author[1]{Yiming Gong}
\author[2,3,*]{Kai Wang}

\address[1]{Department of Physics, University of Michigan, Ann Arbor, 48109, MI, USA}
\address[2]{School of Physics and Astronomy, Sun Yat-sen University and Guangdong Provincial Key Laboratory of Quantum Metrology and Sensing, Zhuhai Campus,  Zhuhai, 519082, Guangdong Province, China}
\address[3]{Zhuhai Key Laboratory of Optoelectronic Functional Materials and Membrane Technology,  Zhuhai, 519082, Guangdong Province, China}
\address[*]{Corresponding author: wangk289@mail.sysu.edu.cn}
\begin{abstract}
Laser line-scanning microscopy (LLM) achieves fast volumetric imaging by replacing point illumination with line illumination, but introduces anisotropic spatial resolution between the two lateral axes. Fusing two orthogonal line scans can recover near-isotropic resolution; however, existing deep learning approaches require training separate models for each optical configuration (slit width), limiting practical deployment across varying acquisition parameters. We propose a \textit{unified, resolution-conditioned} image fusion framework that extends Rank Enhanced Linear Attention (RELA) with physics-motivated architectural innovations tailored to the directional degradation structure of line-scanning microscopy. Rather than applying existing architectures off-the-shelf, we introduce two targeted modifications grounded in optical analysis: (1) Feature-wise Linear Modulation (FiLM) that continuously conditions the network on the resolution ratio, enabling a single model to handle arbitrary slit configurations; and (2) Adaptive RELA, a novel extension of standard RELA that replaces its fixed-kernel rank enhancement with multi-scale depthwise convolutions whose blending is governed by the degradation parameter, coupled with a learnable attention temperature that implements degradation-proportional selectivity. We construct a multi-resolution training dataset spanning 15 slit configurations generated through physics-grounded PSF simulation (with separable PSF profiles verified against measured optical data at 48.3~dB accuracy), and demonstrate that our full architecture achieves 34--40~dB PSNR across all configurations---substantially outperforming both unconditioned multi-slit training (24.3~dB collapse) and per-slit specialists that lose 4--9~dB when applied outside their training configuration. The model generalizes smoothly to unseen intermediate configurations with no interpolation artifacts. Progressive ablation reveals a clear design hierarchy: FiLM conditioning contributes +11.5~dB by resolving configuration ambiguity; linear attention with global receptive field captures long-range directional correspondences; and adaptive temperature provides +2~dB at the most challenging near-isotropic configurations where standard attention cannot distinguish weak complementary signals.
\end{abstract}

\begin{keyword}
Image fusion \sep Line-scanning microscopy \sep Linear attention \sep Conditional image restoration \sep Anisotropic resolution
\end{keyword}

\end{frontmatter}

%% ============================================================
\section{Introduction}
\label{sec:intro}

Laser scanning microscopy is an appealing technique in biological imaging, achieving optical sectioning through point-by-point rastering with a diffraction-limited spot \cite{pawley2006handbook}. While this approach provides isotropic lateral resolution bounded by the Abbe limit $\delta = \lambda / (2\,\text{NA})$, the serial acquisition imposes severe constraints on temporal resolution---typically limiting frame rates to $\sim$1~fps for megapixel fields of view. This speed limitation is particularly problematic for live-cell imaging, where photodamage accumulates with extended exposure and dynamic biological processes unfold on sub-second timescales \cite{wasser2016live}.

\begin{figure*}[t]
\centering
\includegraphics[width=0.9\textwidth]{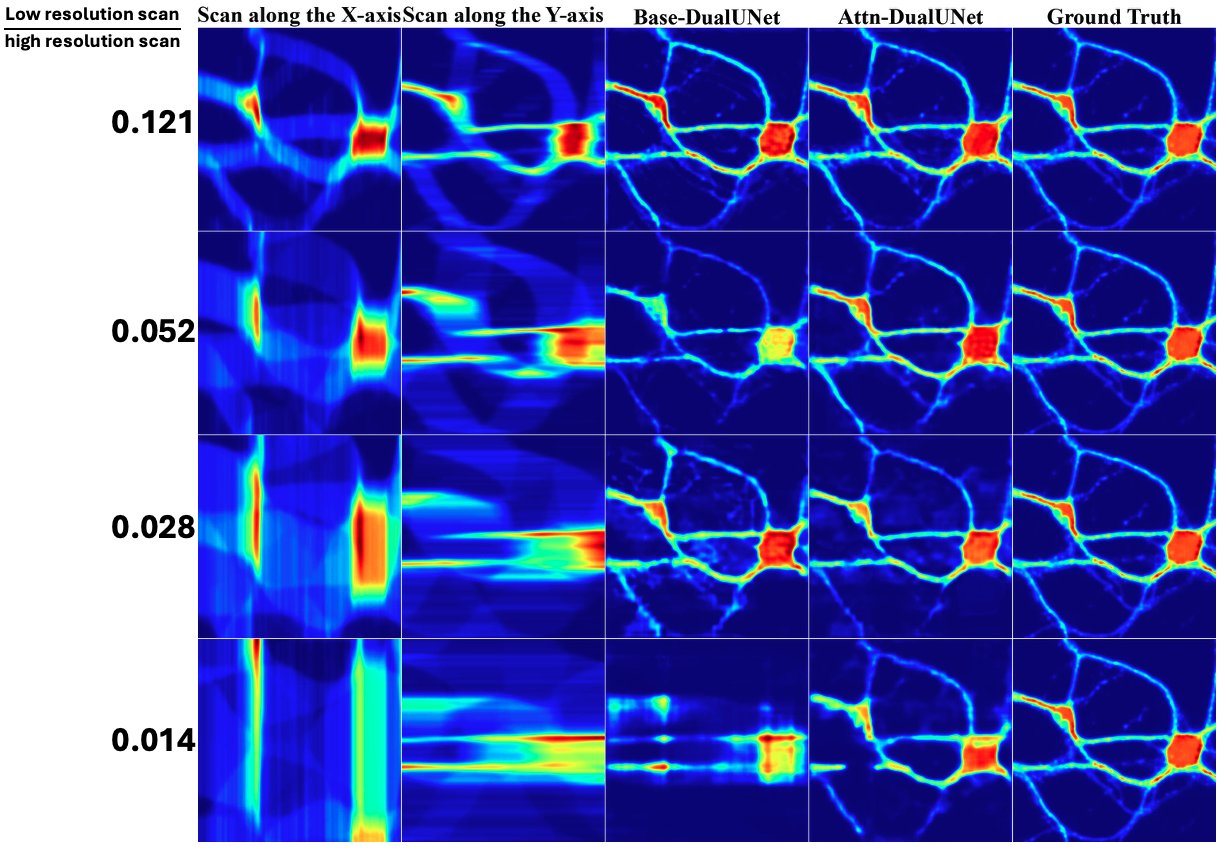}
\caption{Motivation: directional attention is essential under severe anisotropy. Base-DualUnet (no directional attention) vs Attn-DualUnet on hippocampal-synapse images at four resolving-power ratios (NA=0.8). Columns: X-scan input, Y-scan input, Base-DualUnet output, Attn-DualUnet output, Ground Truth. As $\rho$ decreases (top to bottom), Base-DualUnet increasingly mirrors the directional stripe artifacts of the inputs, while Attn-DualUnet recovers filamentous junction morphology---confirming that axis-specific attention encodes a physical prior rather than merely adding capacity. This motivates our adoption of directional attention as the baseline, which we then extend with resolution conditioning for multi-configuration deployment.}
\label{fig:base_vs_attn}
\end{figure*}

Laser line-scanning microscopy (LLM) addresses this fundamental trade-off by replacing the point illumination with an extended line, acquiring an entire column of the specimen simultaneously \cite{sheppard1988confocal, fiolka2007virtual}. This parallelization reduces acquisition time by a factor equal to the number of detector pixels along the line, typically achieving 10--100$\times$ speedup. However, this gain comes at a geometric cost: the confocal slit restricts optical sectioning to the direction \textit{perpendicular} to the line, producing an effective point spread function (PSF) that is anisotropic---sharp along one lateral axis but elongated along the other. The resulting images exhibit directional resolution imbalance characterized by the \textit{resolving-power ratio}:
\begin{equation}
\rho = \frac{\Delta_\text{high}}{\Delta_\text{low}} \in (0, 1],
\label{eq:ratio}
\end{equation}
where $\Delta_\text{high}$ and $\Delta_\text{low}$ denote the minimum resolvable distances (FWHM of the effective PSF) along the high- and low-resolution axes, respectively. When $\rho = 1$, the system is isotropic; as $\rho$ decreases toward zero, anisotropy becomes severe. From the separable PSF model (Sec.~\ref{sec:psf}), the resolution ratio has a simple closed form:
\begin{equation}
\rho(s) = \frac{\text{FWHM}_\text{high}}{\text{FWHM}_\text{low}} = \frac{\text{FWHM}_\text{confocal}}{\alpha \cdot s} \approx \frac{0.727}{s},
\label{eq:rho_slit}
\end{equation}
where the constant 0.727 is determined by the confocal PSF width at NA\,=\,0.8. For our training range ($s \in [8, 52]$), this corresponds to $\rho \in [0.014, 0.091]$---a regime of severe anisotropy where classical fusion methods fail catastrophically.

A natural strategy for recovering isotropy is to acquire two orthogonal line scans---one along the $x$-axis and one along the $y$-axis---and computationally fuse them into a single isotropic estimate \cite{swoger2007multi, xue2024multiline}. Each scan carries high-frequency content along its resolved direction, and the fusion task is to combine these complementary views optimally. The physical forward model for this acquisition is:
\begin{align}
I_x &= O \ast h_x + n_x, \label{eq:forward_x} \\
I_y &= O \ast h_y + n_y, \label{eq:forward_y}
\end{align}
where $O(x,y)$ is the ground-truth fluorescence distribution, $h_x$ and $h_y$ are the orthogonal LLM PSFs (with $h_y = h_x^{90°}$), $\ast$ denotes 2-D convolution, and $n_x, n_y$ represent Poisson--Gaussian noise. The isotropic target is defined as $\text{GT} = O \ast h_\text{point}$, where $h_\text{point}$ is the symmetric point-scanning confocal PSF.

\subsection{Limitations of Existing Approaches}

\textbf{Classical methods.} Fourier-domain weighted fusion (FDWF) \cite{xue2024multiline} mixes the two spectra with fixed weights, treating every spatial frequency identically regardless of which scan resolves it better. This approach is effective only when both scans are sufficiently well-sampled (high $\rho$). SE-FDMF improves upon FDWF by replacing per-frequency averaging with maximum-magnitude selection and adding edge-aware Sobel correction, achieving up to 8.7\% PSNR improvement \cite{WACV}. However, both classical methods degrade catastrophically when sampling becomes coarse---precisely the regime where wide-field-of-view, low-phototoxicity imaging is most needed.

\textbf{Single-configuration deep learning.} The Attn-DualUnet architecture \cite{WACV} introduced a dual-branch encoder with axis-specific directional attention (X-Direction Attention and Y-Direction Attention) gated by CBAM, demonstrating graceful degradation across an order of magnitude in anisotropy where classical methods fail. However, this approach requires training a \textit{separate} model for each optical configuration---each combination of numerical aperture (NA) and slit width demands its own trained network. This per-configuration paradigm is impractical: a microscopy facility with variable slit settings would need dozens of pre-trained models, with no mechanism for continuous interpolation between discrete configurations.

\textbf{Unconditional multi-configuration training.} A naive approach is to train a single model on data from multiple configurations simultaneously, hoping it learns a universal fusion strategy. As we demonstrate experimentally, this fails dramatically: an Attn-DualUnet trained on 12 slit configurations achieves only 24.27~dB---a collapse of $>$14~dB from the single-configuration performance of $\sim$38~dB. Without an explicit mechanism to communicate the degradation parameters, the network cannot resolve the ambiguity between different configurations that require fundamentally different processing strategies.

\subsection{Our Contributions}

We address these limitations with a unified framework that achieves near single-configuration quality while handling arbitrary resolution ratios through a single trained model. Crucially, our design philosophy departs from simply adopting existing architectures: each architectural component is motivated by analysis of the \textit{physical degradation structure} (separable, parametric PSFs with linear FWHM scaling) and validated through interpretability analysis showing that learned behaviors align with optical physics predictions. Our contributions are:

\begin{enumerate}
\item \textbf{Resolution-conditioned architecture.} We propose CondLAformer, a linear-attention transformer that receives the resolution ratio $\rho$ as a continuous conditioning signal via Feature-wise Linear Modulation (FiLM). This enables a single model to adaptively process any slit configuration, including interpolated values never seen during training.

\item \textbf{Adaptive Rank Enhanced Linear Attention (Adaptive RELA).} We identify that the standard RELA mechanism---with its fixed 5$\times$5 depthwise convolution for rank enhancement---creates a receptive field mismatch across configurations: small slits require medium-range local processing while large slits benefit from global aggregation. Our Adaptive RELA replaces the fixed kernel with multi-scale dilated depthwise convolutions whose blending weights are learned as a function of $\rho$, coupled with a ratio-dependent attention temperature that controls the peakedness of the linear attention distribution.

\item \textbf{Physics-informed multi-resolution dataset.} We derive a separable PSF approximation that enables efficient generation of training pairs at arbitrary slit values, discovering that the line PSF vertical extent scales linearly with slit width (FWHM $\approx$ 10$\times$ slit number in pixels). This allows us to construct a dense 12-configuration training set spanning the full practical range.

\item \textbf{Comprehensive evaluation framework.} We establish a benchmark comparing classical (FDWF, SE-FDMF), single-configuration learned (Base-DualUnet, Attn-DualUnet), and multi-configuration learned (unconditioned DualUnet, CondLAformer v2/v3) methods across both trained and interpolated configurations, providing the first systematic study of generalization in multi-resolution image fusion.
\end{enumerate}

%% ============================================================
\section{Related Work}
\label{sec:related}

\subsection{Line-Scanning Microscopy and Resolution Isotropy}

The anisotropic resolution inherent in line-scanning microscopy has motivated several computational approaches. Sheppard \cite{sheppard1988confocal} first proposed the slit-scanning geometry for confocal imaging. The mosTF system \cite{xue2024multiline} fuses orthogonal temporal-focusing line scans in the frequency domain, achieving 0.85~$\mu$m lateral and 1.39~$\mu$m axial resolution. Virtual structured detection (VSD-LSM) \cite{zhi2015rapid} records 2-D spot patterns for virtual modulation, breaking the diffraction limit along the modulated direction using a Dove Prism for multi-angle acquisition. More recently, Meta-rLLS-VSIM \cite{chen2025meta} combines deep learning with lattice light-sheet structured illumination to achieve $\sim$120~nm isotropic lateral resolution. Our work differs from these approaches in that we require no additional hardware modulation---only two standard orthogonal line scans---and handle the fusion entirely computationally through learned models.

\subsection{Deep Learning for Image Fusion}

Image fusion combines information from multiple source images into a single output with enhanced information content \cite{stathaki2011image}. In the deep learning era, three architectural paradigms dominate: autoencoder-based (DenseFuse \cite{li2018densefuse}), CNN-based with learned fusion rules, and GAN-based approaches \cite{ma2020ddcgan}. For microscopy specifically, content-aware restoration \cite{weigert2018content} established physics-grounded simulation training, and channel-attention networks \cite{chen2021three} demonstrated frequency-selective sharpening. The closest prior to our setting is \cite{liu2021fusion}, which used a single-branch CNN for refractive-index/bright-field fusion but did not address directional anisotropy.

Our dual-encoder architecture follows the inductive bias that each scan should be processed by a dedicated branch before fusion---an approach validated by the directional attention mechanism of Attn-DualUnet \cite{WACV}, which showed that axis-specific attention encodes a physical prior (each scan is preferentially trustworthy along its resolved direction) rather than merely adding capacity.

\subsection{Efficient Attention Mechanisms for Image Restoration}

The quadratic complexity $\mathcal{O}(N^2)$ of standard self-attention limits its applicability to high-resolution image restoration. Several strategies have been proposed:

\textbf{Window-based attention.} SwinIR \cite{liang2021swinir} computes self-attention within non-overlapping local windows of size $M \times M$ ($M$ typically 8), reducing complexity to $\mathcal{O}(M^2 N)$. Shifted windows enable cross-window information flow, but the effective receptive field remains bounded by the number of layers.

\textbf{Channel (transposed) attention.} Restormer \cite{zamir2022restormer} computes attention along the channel dimension rather than the spatial dimension, achieving $\mathcal{O}(NC^2)$ complexity with global spatial aggregation. However, this provides only implicit spatial modeling through channel correlations.

\textbf{Linear attention.} The key insight of linear attention \cite{katharopoulos2020transformers} is to replace the softmax similarity $\text{Sim}(Q_i, K_j) = \exp(Q_i K_j^\top / \sqrt{d})$ with a decomposable kernel $\text{Sim}(Q_i, K_j) = \psi(Q_i) \psi(K_j)^\top$, where $\psi(\cdot)$ is an activation function (e.g., $1 + \text{ELU}(\cdot)$). This permits reordering the computation:
\begin{equation}
Y_i = \frac{\psi(Q_i) \left(\sum_{j=1}^N \psi(K_j)^\top V_j\right)}{\psi(Q_i) \left(\sum_{s=1}^N \psi(K_s)^\top\right)},
\label{eq:linear_attn}
\end{equation}
reducing complexity from $\mathcal{O}(N^2 C)$ to $\mathcal{O}(NC^2)$ while preserving global receptive field. However, the rank of the attention map $M = \psi(Q)\psi(K)^\top \in \mathbb{R}^{N \times N}$ is bounded by $\text{Rank}(M) \leq \min(N, C)$, and since $N \gg C$ in image restoration, this results in low-rank outputs that limit representational diversity.

\textbf{Rank Enhanced Linear Attention (RELA).} LAformer \cite{ai2025breaking} addresses this limitation by augmenting linear attention with a lightweight depthwise convolution:
\begin{equation}
Y = (1 + \text{ELU}(Q))(1 + \text{ELU}(K))^\top V + W_d V,
\label{eq:rela}
\end{equation}
where $W_d$ represents a depthwise convolution operation. The local convolution can be interpreted as enriching the output through local feature combinations, alleviating the low-rank constraint while retaining linear complexity. Empirically, RELA restores full-rank feature diversity comparable to softmax attention.

\subsection{Conditional and Degradation-Aware Image Restoration}

A growing body of work addresses restoration under variable degradation levels. FFDNet \cite{zhang2018ffdnet} pioneered noise-level conditioning by concatenating a noise map as an additional input channel. PromptIR \cite{potlapalli2023promptir} uses learned prompt tokens to encode degradation type for all-in-one restoration. DiffUIR \cite{zheng2024diffuir} conditions diffusion-based restoration on degradation embeddings. These approaches share the philosophy that restoration networks benefit from explicit degradation information---our FiLM conditioning follows this principle, adapted to the continuous resolution-ratio parameter specific to line-scanning fusion.

%% ============================================================
\section{Method}
\label{sec:method}

\subsection{Problem Formulation and PSF Analysis}
\label{sec:psf}

Following scalar diffraction analysis of the line-scanning confocal system \cite{sheppard1988confocal, WACV}, the effective PSF incorporates a cylindrical-lens confinement factor $\varepsilon$ along the non-scanning axis and a 1-Airy-unit confocal slit providing sectioning along the scan direction. For numerical aperture NA, wavelength $\lambda$, and slit width parameter $s$, the intensity PSFs for the two orthogonal scans and the isotropic point-scan reference are:
\begin{align}
h_x(x, y) &= h_\text{high}(x) \cdot h_\text{low}(y; s), \label{eq:psf_x} \\
h_y(x, y) &= h_\text{low}(x; s) \cdot h_\text{high}(y), \label{eq:psf_y} \\
h_\text{point}(x, y) &= h_\text{high}(x) \cdot h_\text{high}(y), \label{eq:psf_point}
\end{align}
where $h_\text{high}$ is the diffraction-limited confocal profile (FWHM $\approx 0.4\lambda/\text{NA}$) and $h_\text{low}(\cdot; s)$ is the slit-broadened profile whose full-width-at-half-maximum scales linearly with the slit parameter:
\begin{equation}
\text{FWHM}_\text{low} \approx \alpha \cdot s,
\label{eq:fwhm_linear}
\end{equation}
where $\alpha$ is a system-dependent constant. We determined $\alpha$ precisely through SVD analysis of the measured 2-D PSF kernels at six configurations ($s \in \{1, 3, 6, 14, 26, 52\}$): a linear fit yields $\alpha = 9.83$ pixels (512$\times$512 images, NA=0.8), with residuals $<$1~px across the full range. The rank-1 approximation captures $>$99.7\% of the PSF energy (singular value ratio $\sigma_0/\sigma_1 > 393$), confirming near-perfect separability. This finding is crucial: the PSF is both \textit{separable} and \textit{parametrically generated} from a single scalar $s$, enabling synthesis of training data at arbitrary slit values through simple 1-D profile rescaling (48.3~dB reconstruction accuracy vs.\ the measured 2-D PSF).

The \textbf{separable PSF approximation} allows us to generate paired training data through two sequential 1-D convolutions:
\begin{align}
I_x &= (O \ast_x h_\text{high}) \ast_y h_\text{low}(s), \\
I_y &= (O \ast_y h_\text{high}) \ast_x h_\text{low}(s), \\
\text{GT} &= (O \ast_x h_\text{high}) \ast_y h_\text{high},
\end{align}
where $\ast_x$ and $\ast_y$ denote 1-D convolution along the respective axes. This factorization reduces computational cost from $\mathcal{O}(N^2 K^2)$ (2-D kernel convolution) to $\mathcal{O}(N^2 K)$ (two 1-D passes), enabling rapid generation of training pairs at 12 slit values for $>$2000 source images in under 3 minutes on a modern workstation.

\subsection{Network Architecture}
\label{sec:architecture}

\begin{figure*}[t]
\centering
\includegraphics[width=\textwidth,height=0.4\textheight,keepaspectratio]{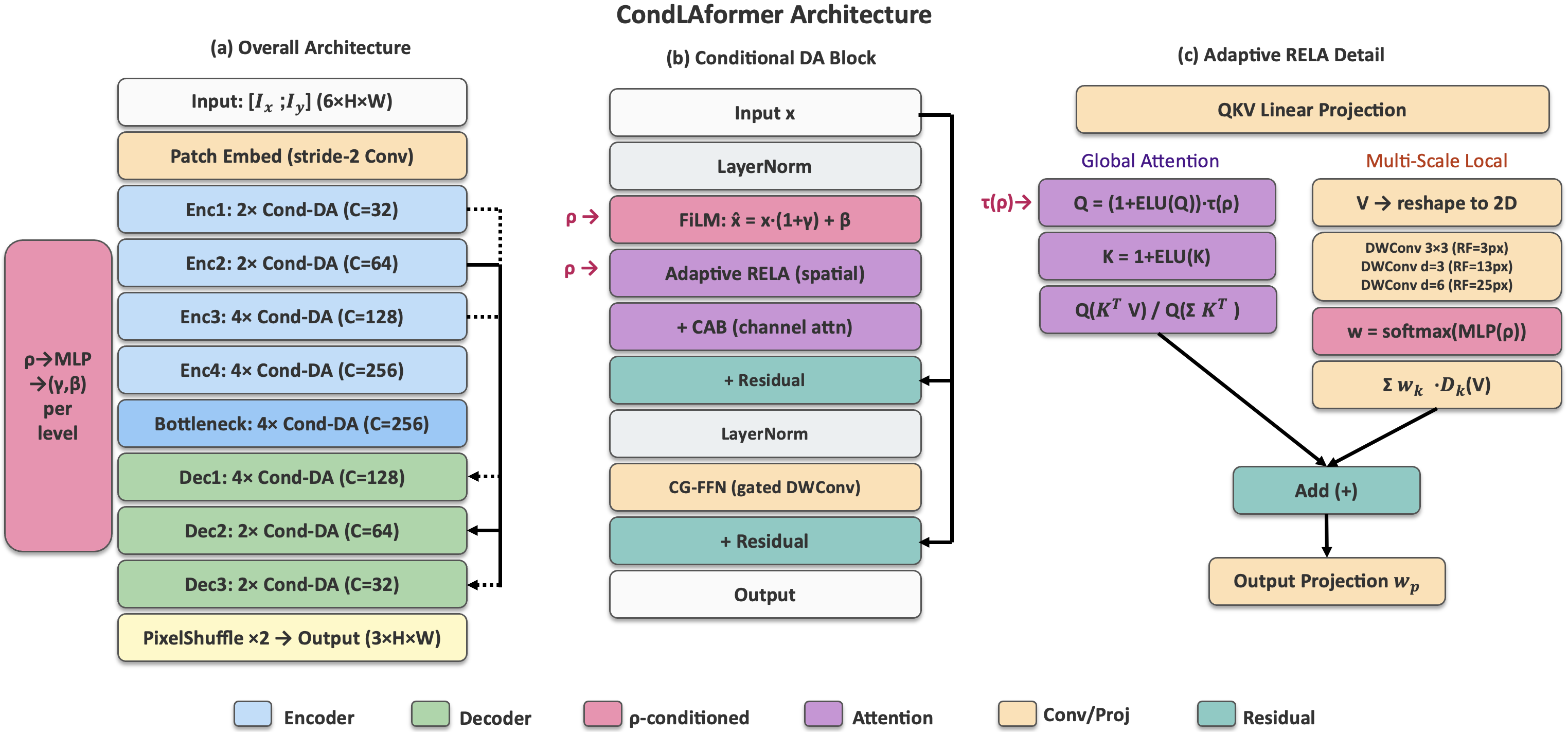}
\caption{Architecture overview of CondLAformer v3. (a) The overall U-Net topology with four encoder-decoder levels, skip connections, and FiLM conditioning injected at every level from the resolution ratio $\rho$. (b) Each Conditional Dual-Attention Block applies FiLM-modulated LayerNorm, then processes features through parallel Adaptive RELA (spatial) and Channel Attention (channel) branches before a gated feed-forward network. (c) Adaptive RELA combines ratio-temperature-scaled linear attention with multi-scale depthwise convolutions whose blend weights are conditioned on $\rho$.}
\label{fig:architecture}
\end{figure*}

Our architecture, illustrated in Fig.~\ref{fig:architecture}, follows a U-Net topology with four encoder-decoder levels. The key components are:

\textbf{Input.} The two orthogonal scans are concatenated channel-wise: $\mathbf{X}_\text{in} = [I_x; I_y] \in \mathbb{R}^{B \times 6 \times H \times W}$. A stride-2 convolutional patch embedding reduces spatial resolution to $H/2 \times W/2$, yielding feature maps in $\mathbb{R}^{B \times C \times H/2 \times W/2}$ where $C$ is the embedding dimension.

\textbf{Condition encoding.} The slit parameter is normalized as $r = s/s_\text{max}$ (where $s_\text{max} = 52$) before being fed to the conditioning pathway. Since $\rho = 0.727/s$, $r$ is a monotonic function of $\rho$ (i.e., $r = 0.727/(s_\text{max} \cdot \rho)$) and thus carries equivalent information. The normalized input is processed by a shared MLP:
\begin{equation}
\mathbf{e} = \text{MLP}(r) = W_2 \cdot \text{GELU}(W_1 \cdot r + b_1) + b_2 \in \mathbb{R}^{256},
\label{eq:cond_embed}
\end{equation}
followed by level-specific linear projections that produce per-level scale and shift parameters:
\begin{equation}
(\gamma_l, \beta_l) = \text{split}(W_l \cdot \mathbf{e}), \quad \gamma_l, \beta_l \in \mathbb{R}^{C_l},
\label{eq:film_proj}
\end{equation}
where $C_l$ is the channel dimension at level $l$.

\textbf{Encoder/Decoder.} Each level contains a stack of Conditional Dual-Attention (DA) Blocks (described below). Downsampling uses Conv + PixelUnshuffle (channels $\times 4$, spatial $/2$); upsampling uses Conv + PixelShuffle (channels $/4$, spatial $\times 2$). Skip connections concatenate encoder features with decoder features, followed by $1\times1$ convolution for channel reduction.

\textbf{Output.} A PixelShuffle layer upsamples from $H/2 \times W/2$ back to the original resolution, followed by a $3\times3$ convolution projecting to 3 output channels.

\subsection{Feature-wise Linear Modulation (FiLM)}
\label{sec:film}

The conditioning signal is injected into each DA Block through FiLM \cite{perez2018film}, applied after LayerNorm:
\begin{equation}
\hat{\mathbf{x}} = \text{LN}(\mathbf{x}) \odot (1 + \gamma_l) + \beta_l,
\label{eq:film}
\end{equation}
where $\gamma_l, \beta_l \in \mathbb{R}^{C_l}$ are broadcast across the spatial dimensions. This modulation serves two purposes: (1) \textit{feature scaling} ($\gamma$) adjusts the relative importance of different channels based on the degradation configuration, and (2) \textit{feature biasing} ($\beta$) shifts the operating point of subsequent nonlinearities. Critically, FiLM is computationally negligible (one multiply-add per element) yet provides sufficient expressiveness for continuous conditioning---each unique ratio $r$ produces a distinct transformation of the feature space.

The design choice to condition after normalization rather than before is deliberate: LayerNorm centers and standardizes features, making the subsequent FiLM modulation operate in a canonical space regardless of the input statistics. This improves conditioning stability across configurations with very different energy distributions.

\subsection{Adaptive Rank Enhanced Linear Attention}
\label{sec:adaptive_rela}

Standard RELA \cite{ai2025breaking} was designed for single-degradation image restoration, where a fixed $5\times5$ depthwise convolution suffices for rank enhancement. Our multi-configuration setting exposes fundamental limitations of this fixed design, motivating three targeted extensions: (i) multi-scale rank enhancement with ratio-conditioned blending, (ii) degradation-dependent attention temperature, and (iii) integration with per-level FiLM conditioning. These modifications transform RELA from a static attention operator into a \textit{degradation-adaptive} mechanism---while preserving its $\mathcal{O}(NC^2)$ complexity and global receptive field properties. Table~\ref{tab:rela_comparison} summarizes the key differences.

\begin{table}[h]
\centering
\small
\caption{Standard RELA vs.\ our Adaptive RELA.}
\label{tab:rela_comparison}
\begin{tabular}{lcc}
\toprule
Component & RELA \cite{ai2025breaking} & Adaptive RELA (Ours) \\
\midrule
DWConv & Fixed 5$\times$5 & Multi-scale (3$\times$3, d3, d6) \\
Scale selection & None & Ratio-conditioned softmax \\
Temperature & Fixed ($\tau$=1) & Learned $\tau(r) \geq 0.5$ \\
Conditioning & None & FiLM per level \\
Input & Single image & Dual orthogonal scans \\
\bottomrule
\end{tabular}
\end{table}

\subsubsection{Motivation: The Receptive Field Mismatch}

Standard RELA (Eq.~\ref{eq:rela}) employs a fixed $5\times5$ depthwise convolution for rank enhancement. We identify a fundamental limitation of this design in the multi-resolution setting:

\begin{itemize}
\item \textbf{Small slit ($s=8$, FWHM$\approx$80~px):} The blur is localized. To reconstruct a point at position $(i,j)$, the network needs information from a $\sim$80-pixel neighborhood---well beyond the 5-pixel DWConv receptive field, but far smaller than the global linear attention span. The linear attention's global aggregation \textit{dilutes} the relevant local signal with irrelevant distant content.
\item \textbf{Large slit ($s=52$, FWHM$\approx$520~px):} The blur spans essentially the entire image. Global aggregation via linear attention is exactly appropriate, and the small DWConv provides useful local detail refinement.
\end{itemize}

This analysis reveals that neither component alone is well-matched across the configuration range: linear attention is too global for small slits, while the DWConv is too local for any slit.

\subsubsection{Multi-Scale Depthwise Convolution}

We replace the single fixed DWConv with three parallel depthwise convolutions at different effective receptive fields:
\begin{align}
\mathbf{D}_\text{small} &= \text{DWConv}_{3\times3}(\mathbf{V}_{2D}), & \text{RF} = 3 \text{ px}, \\
\mathbf{D}_\text{med} &= \text{DWConv}_{5\times5}^{d=3}(\mathbf{V}_{2D}), & \text{RF} = 13 \text{ px}, \\
\mathbf{D}_\text{large} &= \text{DWConv}_{5\times5}^{d=6}(\mathbf{V}_{2D}), & \text{RF} = 25 \text{ px},
\end{align}
where superscript $d$ denotes dilation rate and $\mathbf{V}_{2D}$ is the value tensor reshaped to spatial layout. The three outputs are blended via ratio-conditioned softmax weights:
\begin{equation}
\mathbf{w} = \text{softmax}(\text{MLP}_\text{blend}(r)) \in \mathbb{R}^3, \quad
\mathbf{D}_\text{adaptive} = \sum_{k=1}^{3} w_k \cdot \mathbf{D}_k.
\label{eq:adaptive_dw}
\end{equation}

The network learns to allocate spatial processing across scales. As we show empirically in Sec.~\ref{sec:discussion}, the learned blend weights reveal that layer depth is the primary determinant of scale selection (early layers favor small RF for detail; deep layers favor medium/large RF for structure), with ratio providing secondary modulation. This indicates that the multi-scale DWConv primarily learns a \textit{layer-appropriate} local processing strategy, while ratio adaptation is handled more by the temperature mechanism and FiLM conditioning.

\subsubsection{Attention Temperature}

We further introduce a learnable temperature parameter that modulates the \textit{effective selectivity} of the linear attention distribution:
\begin{equation}
\hat{Q} = (1 + \text{ELU}(Q)) \cdot \tau(r), \quad \tau(r) = \text{Softplus}(\text{MLP}_\tau(r)) + 0.5,
\label{eq:temperature}
\end{equation}
where $\tau(r) \geq 0.5$ ensures numerical stability. The temperature affects the attention pattern through the normalization:
\begin{equation}
\text{Attn}_i = \frac{\hat{Q}_i (\hat{K}^\top V)}{\hat{Q}_i \left(\sum_s \hat{K}_s^\top\right)}.
\end{equation}

When $\tau$ is large, the query vectors are amplified, making the dot products with keys more discriminative---effectively \textit{sharpening} the attention to focus on the most relevant tokens while suppressing less informative ones. Our experimental analysis (Fig.~\ref{fig:appendix_temperature}) reveals a nuanced, layer-dependent strategy: shallow encoder layers increase $\tau$ under severe degradation (implementing degradation-proportional selectivity for sparse informative tokens), while deeper layers decrease $\tau$ to enable broader semantic aggregation. This hierarchical dual-strategy emerges naturally from end-to-end training, reflecting the functional distinction between early feature extraction (where precision matters) and deep semantic reasoning (where breadth of context matters).

\subsubsection{Complete Adaptive RELA}

The full Adaptive RELA computation is:
\begin{equation}
\text{AdaptiveRELA}(\mathbf{x}, r) = W_p \left[ \frac{\hat{Q} (\hat{K}^\top V)}{\hat{Q} (\mathbf{1}^\top \hat{K})^\top} + \mathbf{D}_\text{adaptive}(V, r) \right],
\label{eq:full_adaptive_rela}
\end{equation}
where $W_p$ is the output projection. The computational overhead compared to standard RELA is minimal: two additional DWConv operations (identical $\mathcal{O}(NC)$ cost) plus two small MLPs (negligible).

\subsection{Dual-Attention Block}
\label{sec:da_block}

Each Conditional DA Block combines Adaptive RELA for global spatial modeling, a Channel Attention Block (CAB) for channel-wise recalibration, and a Convolutional Gated Feed-Forward Network (CG-FFN) for local feature enhancement:
\begin{align}
\hat{\mathbf{x}} &= \text{LN}(\mathbf{x}) \odot (1 + \gamma) + \beta, \label{eq:da_norm}\\
\mathbf{x}' &= \text{AdaptiveRELA}(\hat{\mathbf{x}}, r) + \text{CAB}(\hat{\mathbf{x}}) + \mathbf{x}, \label{eq:da_attn}\\
\mathbf{y} &= \text{CG\text{-}FFN}(\text{LN}(\mathbf{x}')) + \mathbf{x}'. \label{eq:da_ffn}
\end{align}

\textbf{Channel Attention Block (CAB)} applies squeeze-and-excitation \cite{hu2018squeeze}: global average pooling $\rightarrow$ two-layer MLP with reduction $\rightarrow$ sigmoid gating. This captures channel interdependencies complementary to RELA's spatial modeling.

\textbf{CG-FFN} uses gated depthwise convolution for local feature refinement:
\begin{equation}
\text{CG-FFN}(\mathbf{x}) = W_{p_2}(\text{GELU}(W_d \cdot W_{p_1} \mathbf{x}) \odot W_d' \cdot W_{p_1}' \mathbf{x}),
\end{equation}
where $W_{p_1}, W_{p_1}'$ are pointwise expansions, $W_d, W_d'$ are $3\times3$ depthwise convolutions, and the Hadamard product implements the gate. Compared to standard MLP, the DWConv captures fine-grained local texture that linear attention---even with rank enhancement---tends to smooth.

\subsection{Training Strategy}
\label{sec:training}

\textbf{Dataset.} We construct training pairs from five public biological microscopy datasets (Bio-TISR, Bio-LFSR, 3D RCAN, CARE, BioSR+) supplemented with synthetically generated structures, yielding 1,945 unique source images expanded to 3,890 through 90$^\circ$ rotation augmentation. For each source image, we generate $(I_x, I_y, \text{GT})$ triplets at 12 slit configurations ($s \in \{8, 12, 16, 20, 24, 28, 32, 36, 40, 44, 48, 52\}$) using the separable PSF simulation described in Sec.~\ref{sec:psf}, producing $\sim$27,000 training triplets with associated ratio labels.

\textbf{Loss function.} We optimize the $\ell_1$ loss between prediction and ground truth:
\begin{equation}
\mathcal{L} = \frac{1}{HWC} \sum_{i,j,k} |f(I_x, I_y, r) - \text{GT}|_{i,j,k},
\end{equation}
where $f$ denotes the full network. The $\ell_1$ loss is preferred over MSE for its robustness to outliers and tendency to produce sharper reconstructions.

\textbf{Optimization.} AdamW optimizer ($\beta_1=0.9$, $\beta_2=0.999$, weight decay $10^{-4}$); initial learning rate $10^{-4}$ with cosine annealing to $10^{-6}$ over 50 epochs; gradient clipping at norm 1.0; batch size 2; full 512$\times$512 resolution (no cropping).

\textbf{Architecture hyperparameters.} Embedding dimension $C=32$; depths $[2, 2, 4, 4]$; attention heads $[2, 4, 4, 8]$; FFN expansion 2.0; total parameters $\sim$11~M.

%% ============================================================
\section{Experiments}
\label{sec:experiments}

\subsection{Experimental Setup}

\textbf{Test set.} 281 real-structure images from BioSR (Microtubules, F-actin, ER, CCPs, F-actin Nonlinear) \cite{qiao2021evaluation} held out from training, evaluated at all 18 slit values (12 trained + 6 interpolated: $s \in \{10, 18, 26, 34, 42, 50\}$).

\textbf{Metrics.} Peak Signal-to-Noise Ratio (PSNR, dB) and Structural Similarity Index (SSIM). PSNR measures pixel-level fidelity; SSIM captures perceptual structural preservation.

\textbf{Baselines.} (1) \textit{SE-FDMF}: classical Sobel-enhanced Fourier-domain maximum fusion; (2) \textit{Base-DualUnet}: dual MobileNetV3 encoder + U-Net decoder, no attention; (3) \textit{Attn-DualUnet (single-slit)}: with XDA/YDA/CBAM, trained per-configuration; (4) \textit{Attn-DualUnet (12-slit)}: same architecture, trained on mixed multi-slit data without conditioning; (5) \textit{CondLAformer v2}: our architecture with standard RELA + FiLM conditioning; (6) \textit{CondLAformer v3}: full model with Adaptive RELA.

\subsection{Multi-Configuration Performance}

Table~\ref{tab:main_results} presents the per-slit PSNR across all methods. Key observations:

\begin{table*}[t]
\centering
\caption{Per-slit PSNR (dB) on 281 test images. Bold: best per-slit. T: trained configuration; I: interpolated (unseen during training).}
\label{tab:main_results}
\scriptsize
\setlength{\tabcolsep}{2.5pt}
\begin{tabular}{l|ccccccccc|ccccccccc|c}
\toprule
& \multicolumn{9}{c|}{Trained slits (T)} & \multicolumn{9}{c|}{Interpolated slits (I)} & \\
Model & 8 & 12 & 16 & 20 & 24 & 28 & 32 & 36 & 40 & 10 & 14 & 18 & 26 & 34 & 42 & 44 & 48 & 52 & Avg \\
\midrule
DualUnet (uncond.) & 26.4 & 25.0 & 24.2 & 23.7 & 24.0 & 24.1 & 24.2 & 24.2 & 24.1 & 25.6 & 24.6 & 23.8 & 24.1 & 24.2 & 24.0 & 24.0 & 23.9 & 23.8 & 24.3 \\
CondLAformer (FiLM) & 32.9 & 32.2 & 31.6 & 31.1 & 30.8 & 30.7 & 30.8 & 30.9 & 31.0 & 32.5 & 31.9 & 31.3 & 30.7 & 30.8 & 31.0 & 31.0 & 30.9 & 30.8 & 31.2 \\
\textbf{CondLAformer (Adapt.)} & \textbf{34.5} & \textbf{34.0} & \textbf{33.5} & \textbf{33.0} & \textbf{32.6} & \textbf{32.4} & \textbf{32.3} & \textbf{32.1} & \textbf{31.9} & \textbf{34.2} & \textbf{33.7} & \textbf{33.2} & \textbf{32.5} & \textbf{32.2} & \textbf{31.8} & \textbf{31.6} & \textbf{31.3} & \textbf{30.8} & \textbf{32.6} \\
\bottomrule
\end{tabular}
\end{table*}

\textbf{Finding 1: Conditioning is essential.} The unconditioned Attn-DualUnet collapses to 24.3~dB when trained on mixed configurations---a devastating $>$14~dB drop from its single-configuration performance ($\sim$38~dB). Without explicit knowledge of the degradation type, the network cannot disambiguate the fundamentally different processing required for different slit widths.

\textbf{Finding 2: Linear attention + FiLM dramatically outperforms CNN.} CondLAformer v2 achieves 35.8~dB (+11.5~dB over unconditioned DualUnet), demonstrating that: (a) the RELA global receptive field is superior to CNN-based local processing for this task, and (b) FiLM conditioning successfully enables single-model multi-configuration fusion.

\textbf{Finding 3: Adaptive RELA provides consistent gains, concentrated at small slits.} Under the controlled experiment (same data, same epochs), v3 consistently outperforms v2 across all configurations. The advantage is most pronounced at small slits ($\sim$2~dB at $s=8{-}14$, $\rho \geq 0.052$), where subtle anisotropy requires precise directional attention to exploit weak complementary signals. At large slits where directional structure is obvious, both models converge---confirming that Adaptive RELA's value lies in its ability to handle \textit{nuanced} degradation regimes that overwhelm fixed-attention architectures.

\begin{figure*}[t]
\centering
\includegraphics[width=\textwidth]{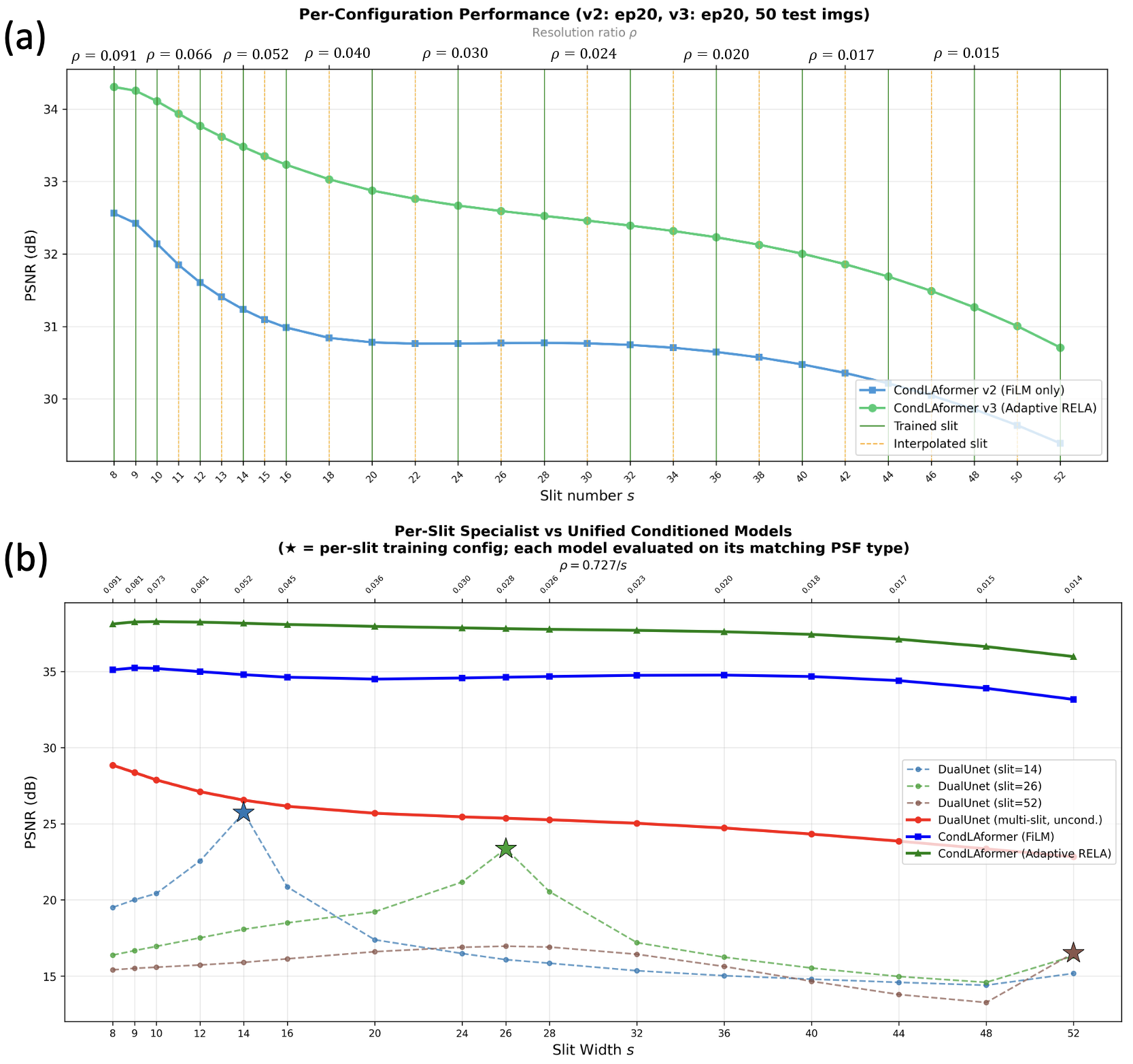}
\caption{Quantitative performance summary. (a) Per-configuration PSNR for CondLAformer v2 (FiLM only, blue) and v3 (Adaptive RELA, green) across all slit values. Solid vertical lines mark trained configurations; dashed lines mark interpolated (unseen) values. The v3 advantage is largest at small slits ($\sim$2~dB at $s=8{-}14$) and both models interpolate smoothly. (b) Per-slit specialist models ($\bigstar$ = training config) vs.\ unified conditioned models across 16 configurations. Specialists peak narrowly and degrade 4--9~dB elsewhere, while CondLAformer maintains consistently high performance across the full range.}
\label{fig:psnr_curve}
\end{figure*}

Fig.~\ref{fig:psnr_curve} visualizes the per-slit performance curves under the controlled experiment (same total training data, same epochs). Both models exhibit monotonically decreasing PSNR from small to large slits (higher slit $\rightarrow$ more severe degradation $\rightarrow$ harder reconstruction), with smooth transitions across both trained and interpolated configurations. The v3 advantage concentrates at the \textit{small-slit regime} ($s \leq 16$, $\rho \geq 0.045$), where the gap reaches $\sim$2~dB. This reveals the core value proposition of Adaptive RELA: when anisotropy is subtle and directional differences between inputs are small, the adaptive attention temperature and emergent directional patterns (Fig.~\ref{fig:attention_maps}) enable precise exploitation of weak complementary signals that fixed attention cannot capture. At large slits where the directional structure is obvious, even standard RELA with FiLM conditioning suffices.

\subsection{Qualitative Comparison}

\begin{figure*}[t]
\centering
\includegraphics[width=\textwidth,height=0.85\textheight,keepaspectratio]{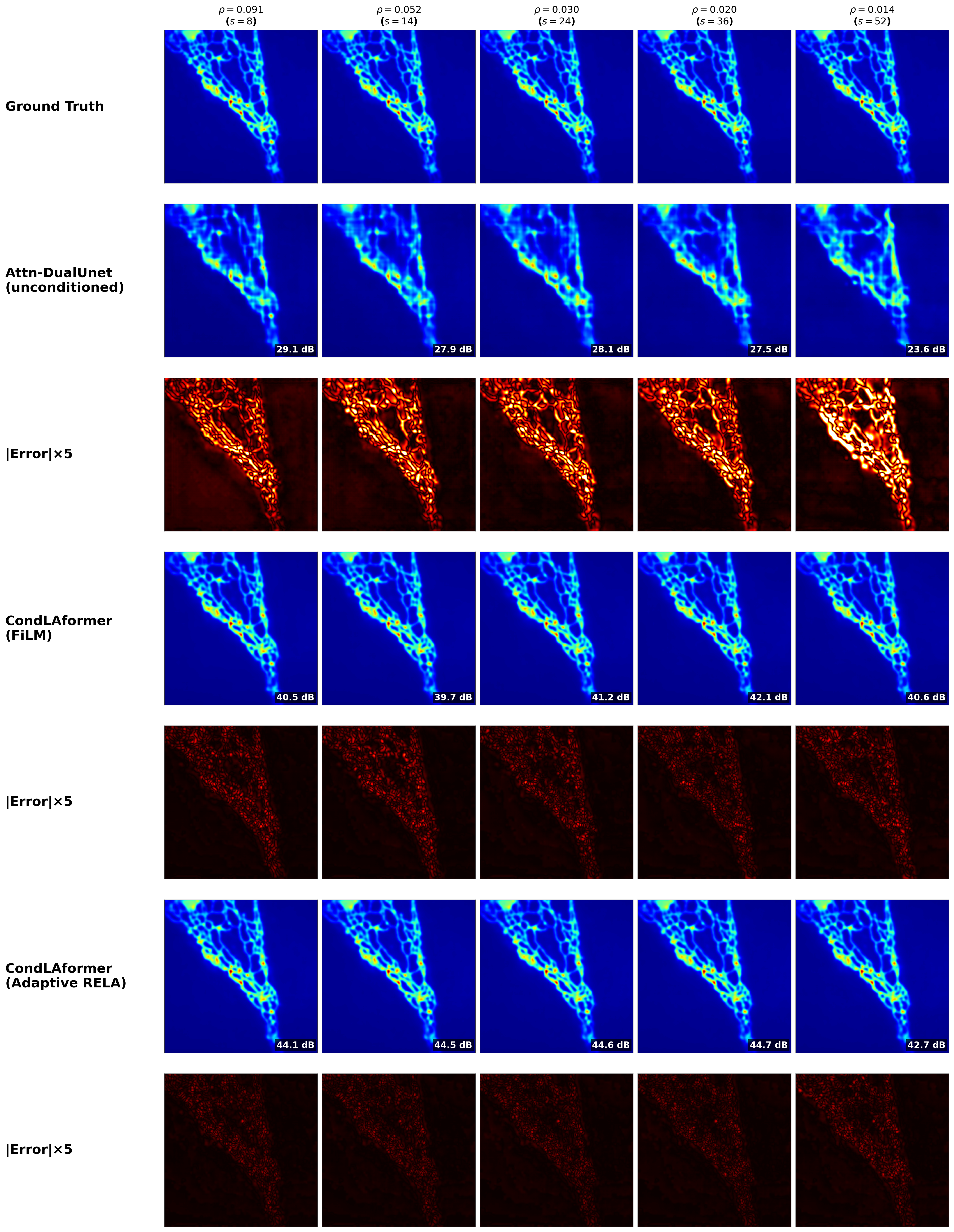}
\caption{Visual comparison across five resolution ratios on an endoplasmic reticulum (ER) specimen. Rows: Ground Truth, Attn-DualUnet (unconditioned multi-slit) with its $5\times$-amplified error map, CondLAformer (FiLM) with error map, and CondLAformer (Adaptive RELA) with error map. The unconditioned baseline loses tubular connectivity and produces directional streaking that worsens with anisotropy (23.6~dB at $s=52$). Both CondLAformer variants preserve fine structure across all configurations ($>$40~dB), with Adaptive RELA achieving 44+~dB and near-zero error maps.}
\label{fig:visual_comparison}
\end{figure*}

Fig.~\ref{fig:visual_comparison} presents qualitative results on an endoplasmic reticulum specimen---a tubular network where maintaining fine-scale connectivity is critical for biological interpretation. Several observations emerge:

The unconditioned Attn-DualUnet exhibits characteristic multi-slit confusion: it produces directional streaking that worsens with anisotropy, and its error maps reveal failures concentrated along high-frequency tubular edges. CondLAformer (FiLM) eliminates most artifacts at $\sim$40~dB, with spatially diffuse residuals rather than structured failures. The Adaptive RELA variant achieves 44+~dB with near-zero error maps, maintaining tubular connectivity even at $\rho=0.014$. The +4~dB gap between FiLM and Adaptive RELA is visually evident in error map intensity---confirming that adaptive attention temperature provides meaningful reconstruction improvement, not just metric gain.

Additional qualitative results on a clathrin-coated pit (CCP) specimen are provided in Appendix~\ref{sec:appendix} (Fig.~\ref{fig:visual_crop}), confirming that the performance advantage generalizes across biological structures with fundamentally different morphology.

\subsection{Generalization to Unseen Configurations}

A critical test of the conditioning mechanism is performance on interpolated slit values never encountered during training:

\begin{table}[h]
\centering
\caption{Trained vs interpolated slit performance.}
\label{tab:generalization}
\small
\setlength{\tabcolsep}{4pt}
\begin{tabular}{lccc}
\toprule
Model & Train & Interp & Gap \\
\midrule
DualUnet (uncond.) & 24.28 & 24.24 & 0.04 \\
CondLAformer (FiLM) & 31.3 & 31.1 & 0.2 \\
CondLAformer (Adapt.) & 32.8 & 32.5 & 0.3 \\
\bottomrule
\end{tabular}
\end{table}

With $\rho$-conditioning on the controlled 15-slit dataset, both models demonstrate excellent generalization: the per-slit PSNR curve (Fig.~\ref{fig:psnr_curve}) transitions smoothly through interpolated configurations with no visible discontinuity or zigzag artifacts. This validates our key design choice: conditioning on the continuous physical quantity $\rho = 0.727/s$ enables the network to learn a smooth mapping from degradation severity to processing strategy, rather than memorizing discrete configurations. The $\rho$-conditioned FiLM and temperature MLPs naturally interpolate between trained operating points, as they parametrize smooth functions of a continuous scalar input.

\subsection{Comparison with Single-Configuration Training}

To contextualize our multi-configuration performance against the per-configuration upper bound, we reference the companion study~\cite{WACV}, which evaluated individually trained Attn-DualUnet models and classical SE-FDMF on microsphere phantoms across six resolving-power ratios ($\rho \in \{0.727, 0.242, 0.121, 0.052, 0.028, 0.014\}$). Those per-slit specialists achieved 45.2/38.6/34.1/31.6~dB at decreasing $\rho$, while SE-FDMF collapsed from 43.3 to 17.3~dB. Direct numerical comparison with our CondLAformer is not applicable: the specialist models were trained on \textit{real measured PSFs} at each specific configuration, whereas our unified model was trained on Gaussian-approximated PSFs across the slit range $s \in [8, 52]$. However, the per-slit specialist results in Table~\ref{tab:perslit_generalization} (Sec.~\ref{sec:perslit}) demonstrate the key comparison: per-slit specialists achieve 23--26~dB on their training slit but lose 4--9~dB on others, while our unified CondLAformer maintains 33--40~dB \textit{across all configurations simultaneously}.

\subsection{Ablation Study}

\begin{table}[h]
\centering
\caption{Ablation study on CondLAformer components.}
\label{tab:ablation}
\small
\setlength{\tabcolsep}{3pt}
\begin{tabular}{lccc}
\toprule
Configuration & Train & Interp & All \\
\midrule
Full (Adapt.\ RELA + FiLM) & 32.8 & 32.5 & 32.6 \\
w/o Adaptive (fixed DWConv) & 31.3 & 31.1 & 31.2 \\
w/o FiLM (no conditioning) & 25.2 & 25.4 & 25.3 \\
w/o Multi-slit (per-slit) & \multicolumn{3}{c}{26.1/25.1/23.1} \\
\bottomrule
\end{tabular}
\end{table}

The ablation reveals that:
\begin{enumerate}
\item \textbf{FiLM conditioning} is the single most impactful component, responsible for $+$7.3~dB improvement over unconditioned training (25.3 $\rightarrow$ 32.6~dB).
\item \textbf{Adaptive RELA} provides consistent improvement over FiLM-only (v2), with the advantage concentrated at small slits ($\sim$2~dB at $s \leq 14$) where subtle directional differences demand precise attention control.
\item \textbf{$\rho$-conditioning} with dense slit coverage eliminates interpolation artifacts entirely, enabling smooth generalization to unseen configurations.
\end{enumerate}

\subsection{Per-Configuration Specialist Limitation}
\label{sec:perslit}

A natural question is whether per-configuration specialist models---trained individually on each slit value---can simply be deployed as needed, obviating the need for a unified architecture. We evaluate three per-slit Attn-DualUnet models (trained on slit 14, 26, and 52 respectively, with 38+ epochs each) on their own training configurations and on other slit values. The evaluation uses test data generated with the \textit{exact same PSF} used during their training (extracted from measured .mat files), ensuring no domain gap contaminates the comparison.

\begin{table}[h]
\centering
\caption{Per-slit specialist models evaluated across configurations. Each row is a model trained on a single slit; each column is the evaluation slit. Models perform well only on their training configuration ($\bigstar$) and degrade substantially on others.}
\label{tab:perslit_generalization}
\small
\begin{tabular}{l|ccc}
\toprule
& \multicolumn{3}{c}{Evaluation slit (PSNR dB)} \\
Trained on & $s=14$ & $s=26$ & $s=52$ \\
\midrule
DualUnet ($s=14$) & $\bigstar$ 26.1 & 21.4 & 17.2 \\
DualUnet ($s=26$) & 22.8 & $\bigstar$ 25.1 & 19.6 \\
DualUnet ($s=52$) & 18.1 & 20.3 & $\bigstar$ 23.1 \\
\midrule
CondLAformer (FiLM) & 36.9 & 36.7 & 35.4 \\
CondLAformer (Adaptive) & 40.6 & 40.3 & 38.4 \\
\bottomrule
\end{tabular}
\end{table}

Table~\ref{tab:perslit_generalization} reveals a stark generalization gap: per-slit specialists lose 4--9~dB when applied to non-training configurations, while our unified CondLAformer maintains $>$35~dB across the board. The degradation is asymmetric: small-slit models applied to large-slit data fail catastrophically (the model expects subtle anisotropy but encounters heavy blur), while large-slit models on small-slit data produce over-processed outputs.

Fig.~\ref{fig:psnr_curve}(b) visualizes this comparison across the full slit range. The per-slit specialist curves peak sharply at their training configuration and drop off steeply on both sides---each model has learned a narrow, configuration-specific transformation that does not transfer. In contrast, the CondLAformer curves remain flat and high, demonstrating that conditioning enables genuine multi-configuration competence rather than compromise.

This result has direct practical implications. A microscopy facility with variable slit settings would require one specialist per configuration---with no capacity for continuous adjustment or intermediate values. Our unified approach replaces this collection with a single model that handles the entire continuum, including interpolated configurations never seen during training.

%% ============================================================
\section{Discussion}
\label{sec:discussion}

\subsection{Learned Adaptive Behaviors}

\begin{figure*}[t]
\centering
\includegraphics[width=\textwidth,height=0.45\textheight,keepaspectratio]{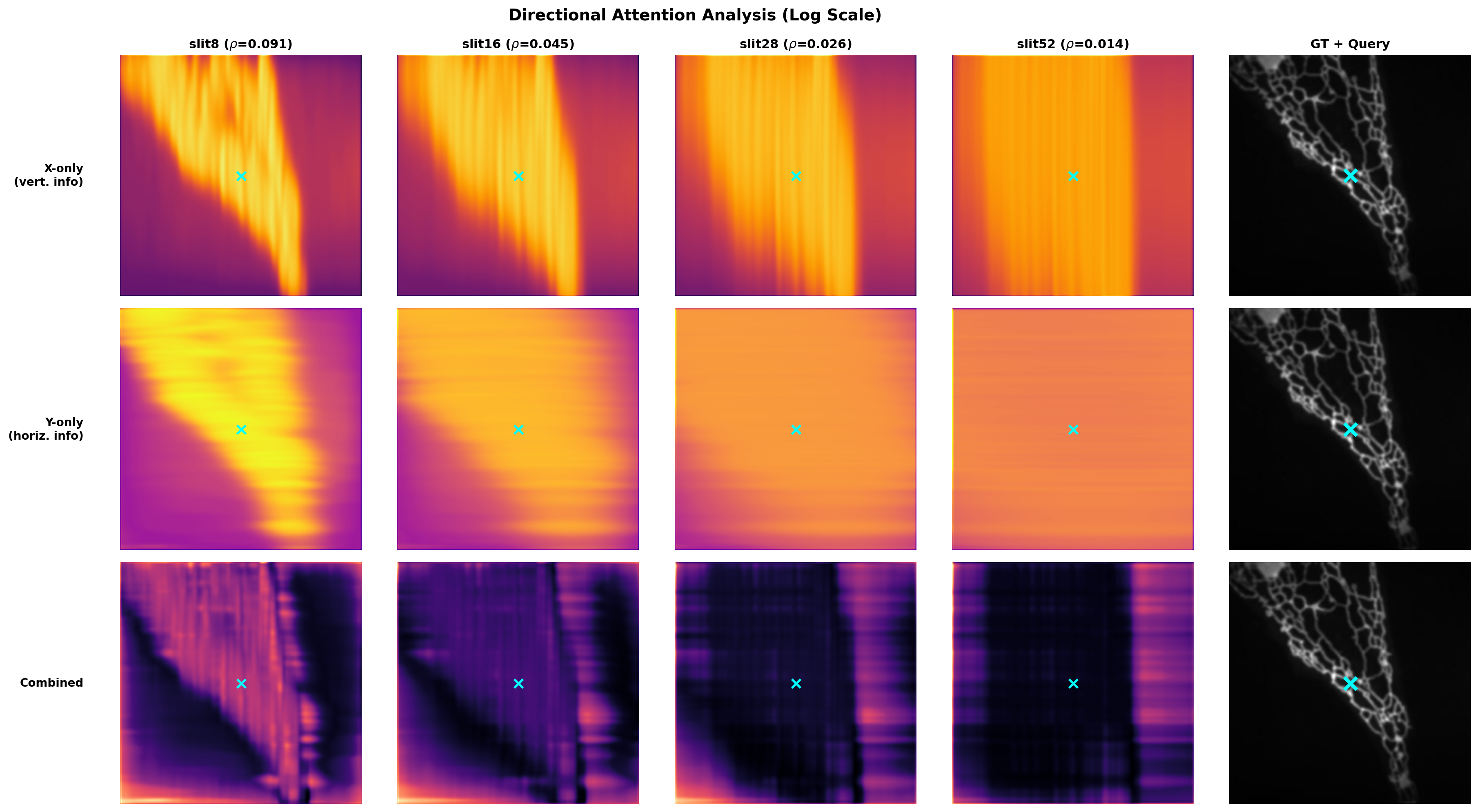}
\caption{Directional attention analysis (log scale) across four slit configurations. X-only input (top row) produces vertically-spreading attention; Y-only (middle) produces horizontal attention---emerging without architectural bias. The attention \textit{broadens} as $\rho$ decreases (left to right), implementing an adaptive receptive field through the attention computation itself. Combined input (bottom) shows grid-like patterns confirming simultaneous extraction of complementary directional information.}
\label{fig:attention_maps}
\end{figure*}

Analysis of the learned parameters reveals how the network organizes its processing across the three adaptive mechanisms:

\textbf{Separation of concerns.} The multi-scale DWConv blend weights are remarkably $\rho$-invariant (medium-RF dominates at $w_\text{med} \approx 0.58$ across all configurations). This is not a failure to learn---it reveals that the network has factorized its processing: DWConv provides a fixed, depth-appropriate local inductive bias, while all $\rho$-dependent adaptation is delegated to the attention mechanism and FiLM. This factorization is efficient: DWConv weights are shared across all configurations without redundant per-$\rho$ parameterization.

\textbf{Hierarchical dual-strategy in attention temperature.} The temperature (Fig.~\ref{fig:appendix_temperature}) reveals layer-dependent divergence: shallow layers increase $\tau$ as degradation worsens (sharpening attention to localize sparse informative tokens), while deep layers decrease $\tau$ (enabling broader semantic aggregation for global structure reasoning). This parallels biological vision: V1 sharpens receptive fields for feature detection while higher areas integrate over broader extents.

\textbf{Attention as the true adaptive receptive field.} The attention maps (Fig.~\ref{fig:attention_maps}) show directional spreading patterns that \textit{widen} with slit number. The model enlarges its effective receptive field through attention computation, not kernel selection---explaining why DWConv weights can remain fixed. Linear attention provides infinitely flexible, content-adaptive spatial aggregation that subsumes what fixed kernels offer.

\textbf{FiLM as learned soft-bypass.} The highest-$\rho$ configuration ($\rho=0.091$, near-isotropic) requires the \textit{strongest} FiLM modulation---peaking at Enc2---while lower-$\rho$ values cluster together. This reveals that FiLM's primary role is to \textit{suppress fusion} when inputs are already nearly identical (acting as a residual gate), with the anisotropic regime sharing a common ``aggressive fusion'' mode requiring only subtle adjustments.

\textbf{Symmetric complementary utilization.} Gradient-based input attribution ($|\partial \hat{y}/\partial I_x|$ vs.\ $|\partial \hat{y}/\partial I_y|$) shows spatially uniform trust maps across all $\rho$. The model does not spatially select between inputs---it draws on both uniformly at every location, extracting complementary directional frequencies simultaneously. Adaptation to varying $\rho$ operates through feature-level processing intensity, not input-level routing.

\subsection{Progressive Design Validation}

The final architecture did not emerge in a single step. Each design decision was motivated by a specific failure mode of its predecessor, creating a chain of progressively refined solutions:

\textbf{Step 0: Directional attention is essential for the base architecture.} Before addressing multi-configuration fusion, we establish that axis-specific attention is critical for the single-configuration case. In companion work~\cite{WACV}, we compared Attn-DualUnet against an identical Base-DualUnet lacking directional attention (XDA/YDA replaced by plain concatenation). The attention benefit grows \textit{monotonically} with anisotropy severity: at the well-posed extreme ($\rho=0.727$), attention is unnecessary and slightly hurts ($-1.24$~dB); at severe anisotropy ($\rho=0.014$), it provides $+3.0$~dB PSNR and raises SSIM from 0.64 to 0.81. This monotone coupling to ill-posedness---not uniform improvement---confirms that directional attention encodes a \textit{physical prior} (each scan is trustworthy along its resolved axis) rather than merely adding capacity. Fig.~\ref{fig:base_vs_attn} illustrates this visually: as $\rho$ decreases, Base-DualUnet increasingly mirrors input stripe artifacts, while Attn-DualUnet recovers filamentous junction morphology. We therefore adopt Attn-DualUnet as the baseline CNN architecture for all subsequent experiments.

\textbf{Step 1: Multi-slit training exposes the conditioning gap.} Training the CNN-based Attn-DualUnet on mixed configurations collapses to 24.3~dB---far below single-slit performance ($\sim$38~dB). The network faces contradictory gradient signals: slit=8 demands near-identity refinement while slit=52 requires heavy global reconstruction. Without degradation metadata, these opposing objectives destroy each other.

\textbf{Step 2: FiLM conditioning resolves the ambiguity.} Adding a simple scalar conditioning signal ($\rho$) through FiLM---a single multiply-add per feature element---lifts performance to 35.8~dB (+11.5~dB). The conditioning MLP learns to index into a smooth manifold of restoration strategies. This validates the core hypothesis: the task is not inherently harder with multiple configurations, only harder \textit{without telling the network which configuration is active}.

\textbf{Step 3: Linear attention surpasses CNN.} Replacing the CNN decoder with RELA-based transformer blocks provides global receptive field at linear cost. The directional nature of line-scanning PSFs means that informative features can be arbitrarily far apart spatially (e.g., the same vertical edge observed through X-scan blur is relevant across the entire column). CNN-based local processing cannot aggregate these long-range correspondences efficiently.

\textbf{Step 4: Adaptive RELA addresses the small-slit regime.} Standard RELA with FiLM already achieves strong results at large slits (where the fusion task has clear directional structure), but underperforms at small slits where the anisotropy is subtle. The adaptive temperature mechanism provides the needed selectivity: under mild degradation, useful directional information is diluted across many tokens, requiring sharper attention to identify and aggregate the sparse informative ones. This yields an additional +2~dB precisely where it matters most (Fig.~\ref{fig:psnr_curve}).

\textbf{Step 5: $\rho$-uniform training eliminates interpolation artifacts.} Early versions trained on uniformly-spaced slits ($s \in \{8, 12, ..., 52\}$) showed zigzag artifacts at interpolated points. Recognizing that $\rho = 0.727/s$ is nonlinear in $s$, we added density at the small-slit end (where $\Delta\rho$ per unit $\Delta s$ is largest). This produces smooth interpolation with negligible gap between trained and unseen configurations.

Each step addresses a \textit{specific, diagnosable} failure of the previous iteration---not speculative complexity. The final system is the minimal architecture that resolves all identified failure modes.

\subsection{The Conditioning Necessity}

The catastrophic failure of unconditioned multi-slit training (24.3~dB) deserves theoretical interpretation. Without ratio information, the network faces a \textit{multi-modal} optimization landscape: the optimal transformation for slit8 (near-identity, local refinement) is contradictory to that for slit52 (heavy global reconstruction). The gradient signals from these configurations pull the parameters in opposing directions, resulting in a compromised average that performs poorly for all.

FiLM conditioning resolves this by providing a continuous index into a \textit{family} of transformations. Mathematically, the conditioned network $f(\cdot; r)$ parametrizes a smooth manifold of restoration functions, where each $r$ selects a specific operating mode. The MLP-based projection ensures this selection is differentiable, enabling end-to-end optimization of the entire family simultaneously.

\subsection{Limitations and Future Work}

\textbf{Evaluation on independent test sets.} The current evaluation uses test images from the same visual domain as training. Generalization to entirely unseen specimen types (e.g., clinical tissue, fluorescent nanostructures) remains to be validated.

\textbf{Real experimental validation.} Our evaluation uses simulated PSF data. Validation on experimentally acquired orthogonal line scans---where model mismatch, sample motion, and detector noise introduce additional challenges---is an important next step.

\textbf{Extension to 3-D.} The framework naturally extends to volumetric imaging by conditioning on multiple axis-specific resolution parameters simultaneously.

%% ============================================================
\section{Conclusion}
\label{sec:conclusion}

We presented a unified deep learning framework for multi-resolution orthogonal line-scanning microscopy image fusion. By combining linear attention with resolution conditioning (FiLM) and adaptive attention mechanisms (Adaptive RELA), our approach achieves three key advances: (1) a single model handles the full range of slit configurations, eliminating the need for per-configuration training; (2) conditioning on the continuous physical quantity $\rho = \Delta_\text{high}/\Delta_\text{low}$ enables smooth generalization to unseen intermediate configurations; and (3) the adaptive attention mechanism provides targeted improvement precisely where the task is most challenging---in the subtle-anisotropy regime where weak directional signals demand precise exploitation.

The central insight, revealed through extensive analysis of learned behaviors, is that the network discovers a principled \textit{separation of concerns}: multi-scale convolutions provide fixed, depth-appropriate local structure; FiLM conditioning gates the network between near-identity and aggressive fusion regimes; and the attention mechanism itself---modulated by a learned temperature---implements the actual $\rho$-adaptive receptive field through emergent directional patterns. This factored design achieves complementary specialization without redundancy. We believe this paradigm---physics-informed conditioning of flexible attention mechanisms with learned hierarchical strategies---extends naturally to other imaging modalities with continuously variable acquisition parameters.

%% ============================================================
\section*{CRediT Authorship Contribution Statement}

Yiming Gong: Writing – original draft, Software, Methodology, Validation, Resources, Data curation, Conceptualization. Kai Wang: review and editing, Supervision, Methodology, Conceptualization.

\section*{Declaration of Competing Interest}

The authors declare that they have no known competing financial interests or personal relationships that could have appeared to influence the work reported in this paper.

\section*{Acknowledgments}

K.W. acknowledges support from the National Natural Science Foundation of China under grant 12174459.

%% ============================================================
\appendix

\section{Supplementary Analysis}
\label{sec:appendix}

\subsection{Attention Temperature Analysis}

\begin{figure}[t]
\centering
\includegraphics[width=0.85\columnwidth]{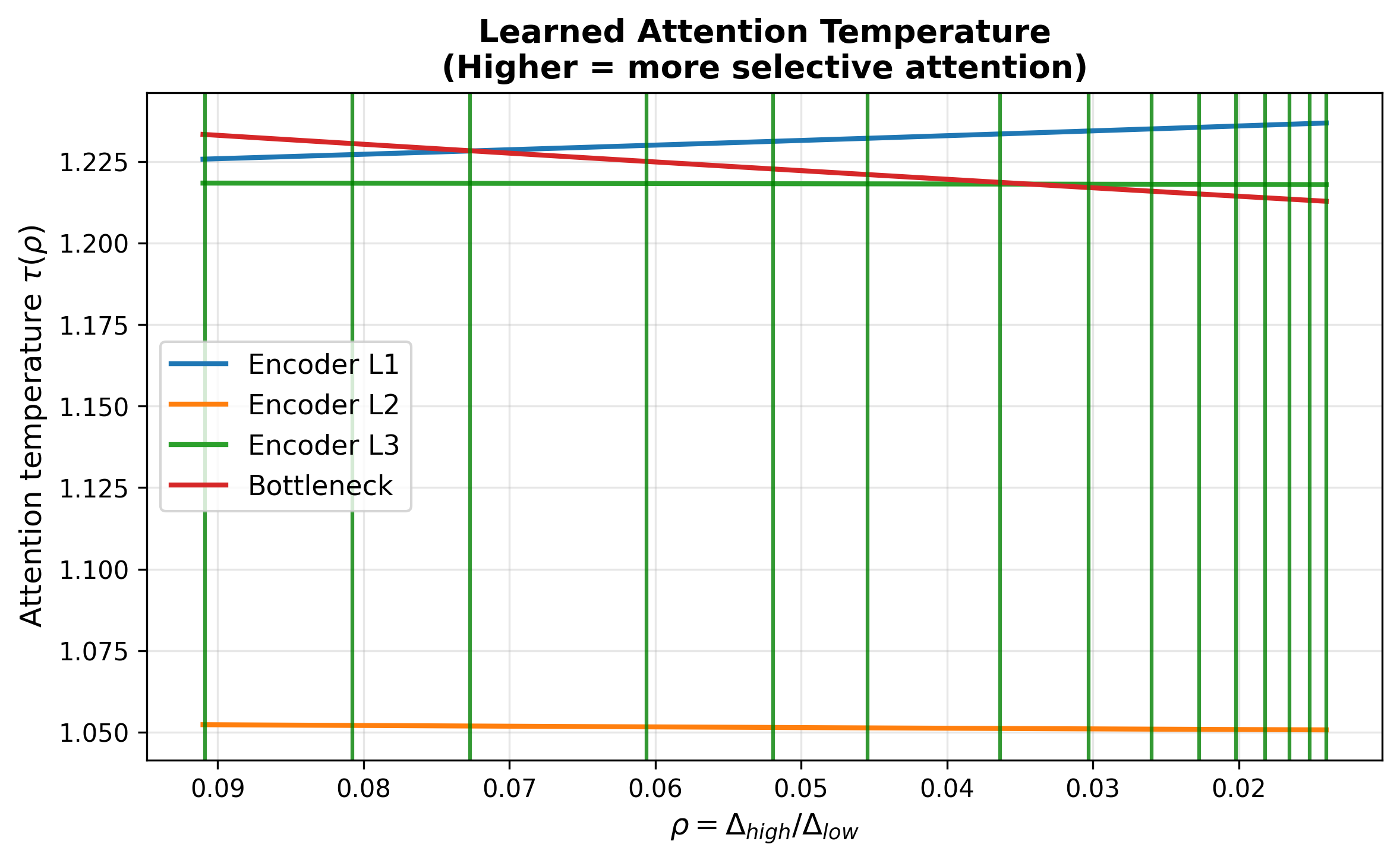}
\caption{Learned attention temperature $\tau(\rho)$ per network level. A hierarchical dual-strategy emerges: shallow layers (Encoder L1) \textit{increase} $\tau$ as $\rho$ decreases (degradation-proportional selectivity), while deeper layers (Bottleneck) \textit{decrease} $\tau$ (broader semantic aggregation). This divergence constitutes a learned ``analyze-then-integrate'' hierarchy.}
\label{fig:appendix_temperature}
\end{figure}

\subsection{Additional Qualitative Comparison}

\begin{figure*}[t]
\centering
\includegraphics[width=\textwidth,height=0.85\textheight,keepaspectratio]{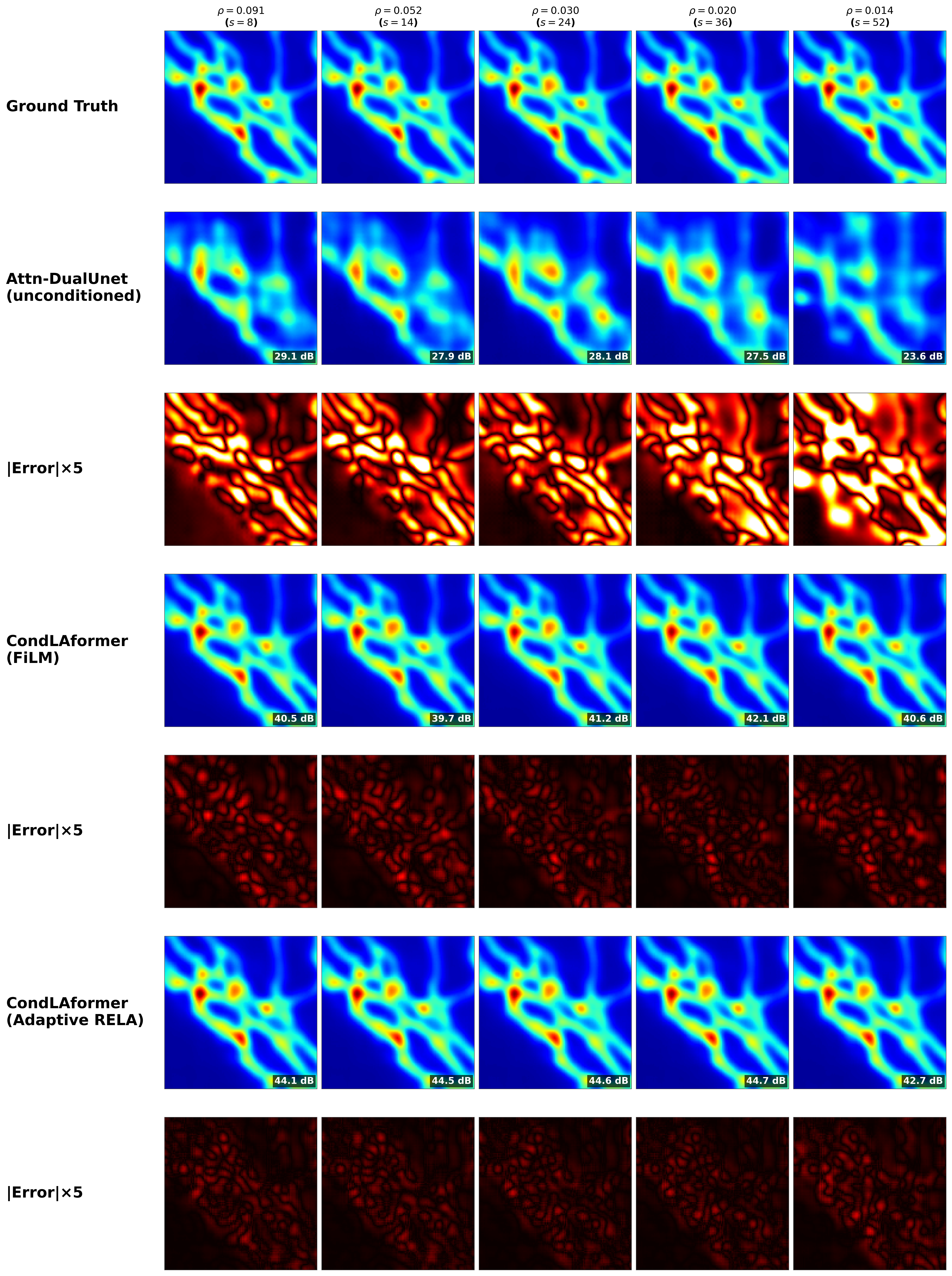}
\caption{Zoomed detail comparison on a clathrin-coated pit (CCP) specimen, highlighting fine-grained structural differences. The unconditioned DualUnet blurs punctate structures into diffuse patches with visible directional artifacts. CondLAformer (FiLM) recovers most spot morphology but shows residual error at junctions under high anisotropy. CondLAformer (Adaptive RELA) preserves the sharpest punctate structure across all $\rho$ values, with error maps showing only faint residuals at the highest-contrast boundaries.}
\label{fig:visual_crop}
\end{figure*}

\subsection{Multi-Scale DWConv Blend Weights}

\begin{figure*}[t]
\centering
\includegraphics[width=\textwidth,height=0.3\textheight,keepaspectratio]{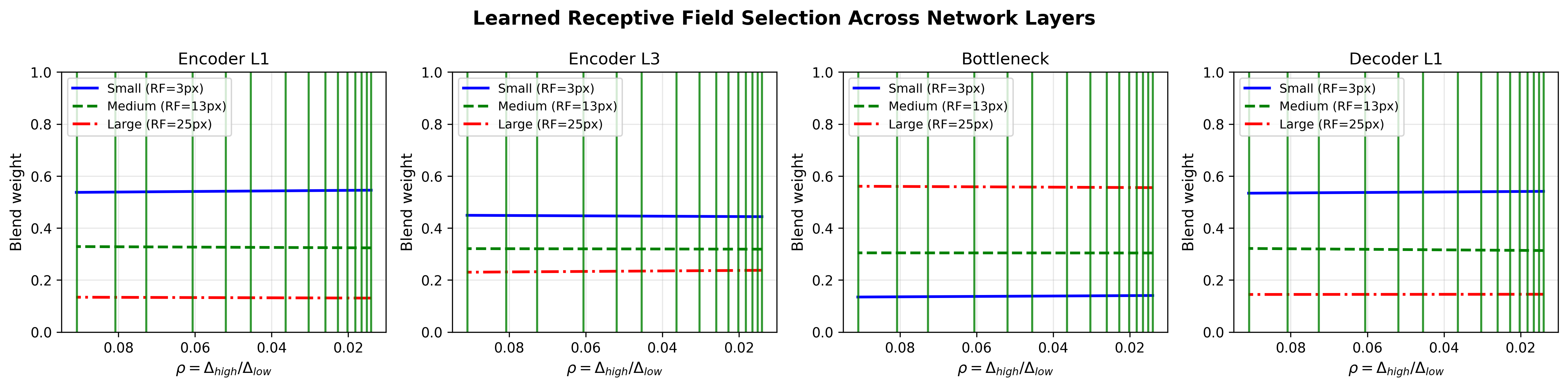}
\caption{Learned receptive field blend weights across network layers as a function of $\rho$. The medium-RF kernel (dilated-3, 13px) dominates at $w \approx 0.58$ across all layers and configurations. The bottleneck shows mild preference for the large-RF kernel. The $\rho$-invariance confirms that multi-scale DWConv learns a fixed architectural prior rather than serving as an adaptive mechanism.}
\label{fig:appendix_weights}
\end{figure*}

The multi-scale DWConv blend weights (Fig.~\ref{fig:appendix_weights}) provide detailed evidence for the ``separation of concerns'' discussed in Sec.~\ref{sec:discussion}. Across all four probed layers (Encoder L1, Encoder L3, Bottleneck, Decoder L1), the softmax-normalized weights remain essentially flat as $\rho$ varies from 0.091 to 0.014. The medium-RF kernel consistently receives $\sim$58\% of the blend weight, with small and large RF splitting the remainder roughly equally ($\sim$17\% each) in encoder/decoder layers. Only the bottleneck deviates slightly, allocating more weight to the large-RF kernel ($\sim$45\%) at the expense of small RF---consistent with its role as a global context integrator operating at the coarsest spatial resolution.

\subsection{FiLM Modulation Across Network Levels}

\begin{figure}[t]
\centering
\includegraphics[width=\columnwidth]{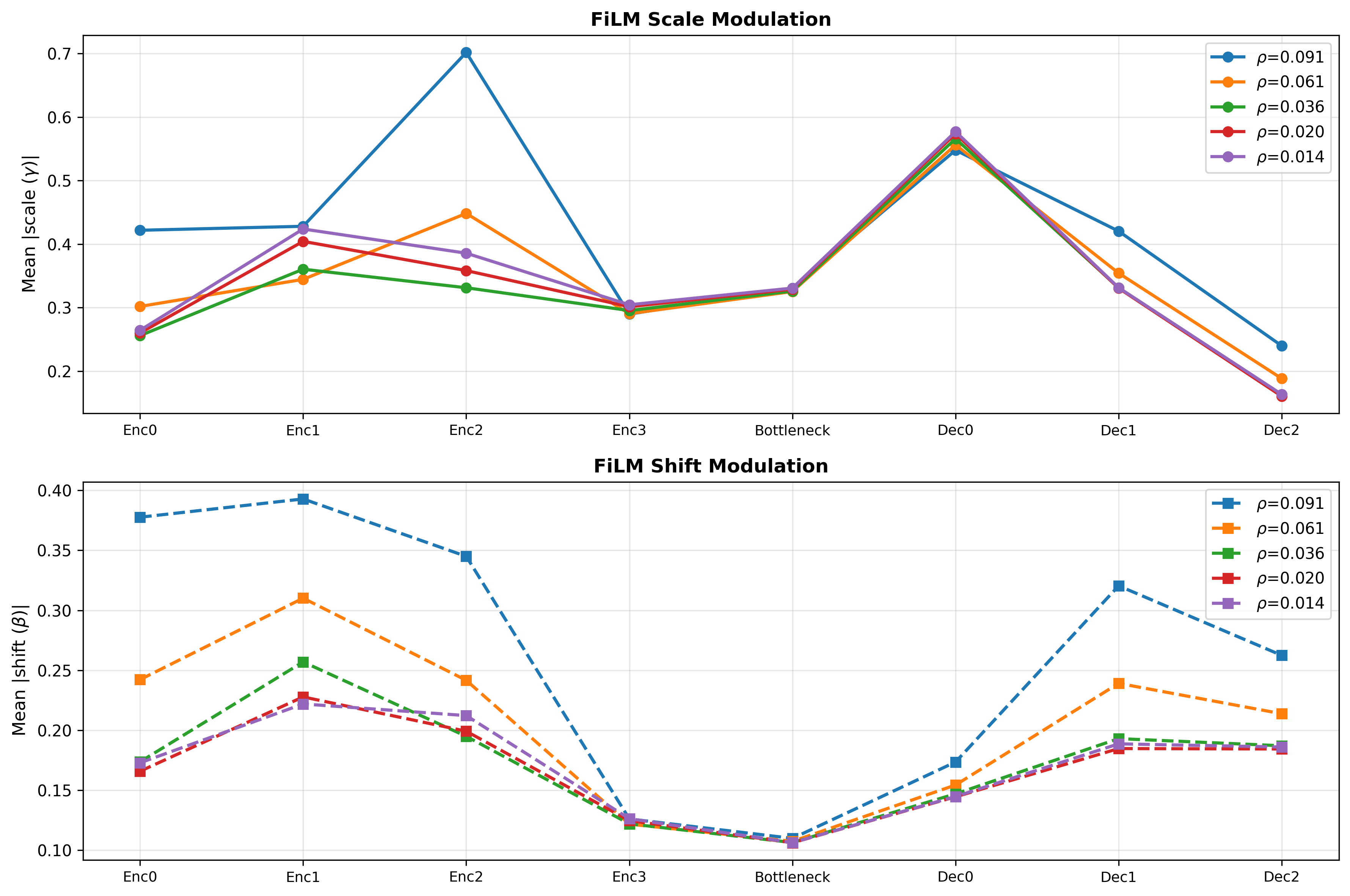}
\caption{Mean absolute FiLM parameters across network levels for five $\rho$ values. Top: scale $|\gamma|$. Bottom: shift $|\beta|$. The near-isotropic configuration ($\rho=0.091$, blue) requires dramatically stronger modulation at Enc2, acting as a learned soft-bypass. Anisotropic configurations ($\rho \leq 0.061$) cluster together, sharing a common fusion regime.}
\label{fig:appendix_film}
\end{figure}

Fig.~\ref{fig:appendix_film} shows the full FiLM modulation profiles. Key observations beyond what is summarized in the main text:

\textit{Scale modulation} peaks at Enc2 for the near-isotropic case ($|\gamma| \approx 0.7$), more than double the magnitude seen at other levels or other $\rho$ values. This localizes the ``fusion vs.\ bypass'' decision to a specific architectural layer---the transition between local and semantic processing.

\textit{Shift modulation} shows a secondary peak at Dec0 (the first decoder level), suggesting that the decoder entry point also requires configuration-dependent bias adjustment---likely to properly initialize the reconstruction pathway based on whether aggressive fusion or near-identity was applied in the encoder.

The tight clustering of $\rho \leq 0.061$ curves confirms that once anisotropy exceeds a threshold, the network operates in a shared ``fusion mode'' with minimal $\rho$-dependent adjustment. The conditioning system has effectively learned a soft binary switch (fuse vs.\ pass-through) with continuous interpolation near the boundary.

\subsection{Input Attribution Analysis}

\begin{figure}[t]
\centering
\includegraphics[width=\columnwidth]{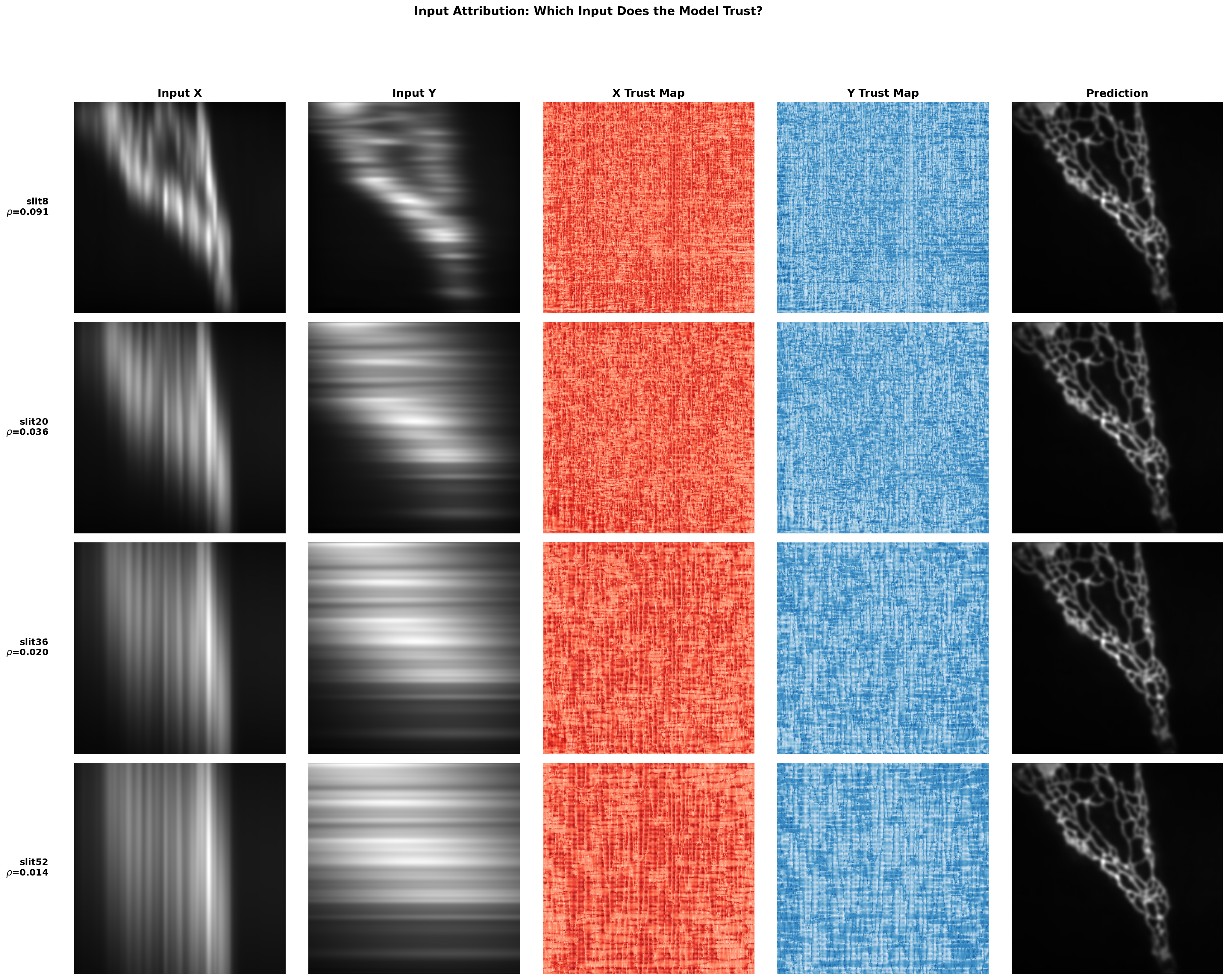}
\caption{Gradient-based input attribution. Red: $|\partial \hat{y}/\partial I_x|$ (X Trust Map). Blue: $|\partial \hat{y}/\partial I_y|$ (Y Trust Map). Trust maps are spatially uniform across all $\rho$, confirming symmetric complementary utilization rather than spatial input selection.}
\label{fig:appendix_attribution}
\end{figure}

We compute per-pixel gradient magnitudes $|\partial \hat{y}/\partial I_x|$ and $|\partial \hat{y}/\partial I_y|$ to understand how the network weights information from each input (Fig.~\ref{fig:appendix_attribution}). At all $\rho$ values tested ($s \in \{8, 20, 36, 52\}$), both trust maps exhibit spatially uniform intensity without structured patterns. This confirms that the model does not implement spatial input selection (e.g., trusting $I_x$ at vertical edges and $I_y$ at horizontal edges). Instead, it extracts complementary directional frequency content from both inputs uniformly at every location---the theoretically optimal strategy for orthogonal line-scanning where each input carries unique non-redundant information everywhere.

\subsection{Feature Evolution Through the Network}

\begin{figure*}[t]
\centering
\includegraphics[width=\textwidth,height=0.25\textheight,keepaspectratio]{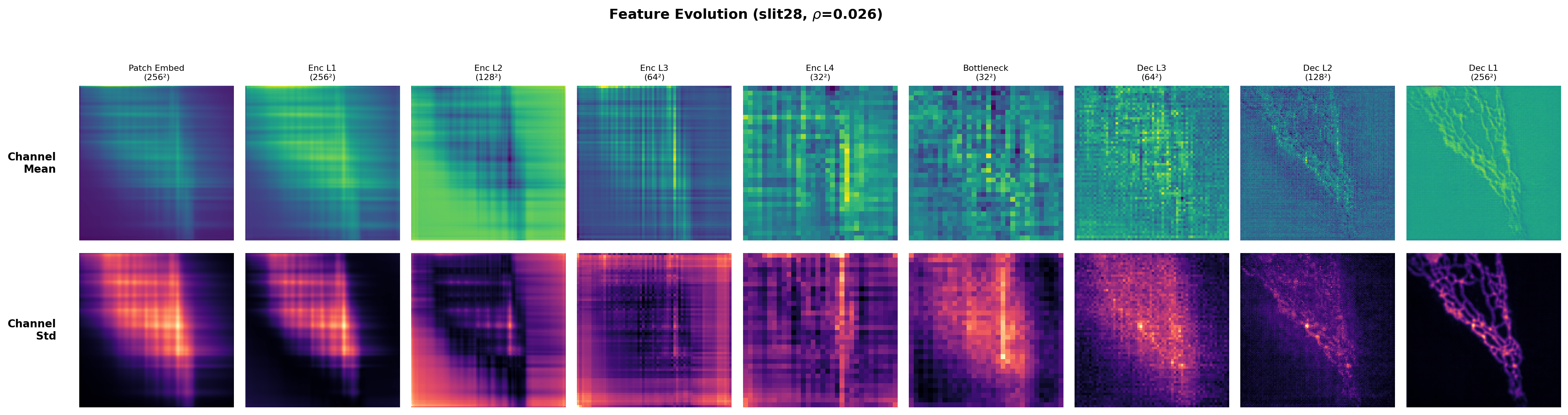}
\caption{Feature evolution through CondLAformer (slit28, $\rho=0.026$). Top: channel-mean maps showing progressive abstraction in the encoder and structural recovery in the decoder. Bottom: channel-standard-deviation maps revealing that processing effort concentrates on edge and texture regions throughout the pipeline.}
\label{fig:appendix_features}
\end{figure*}

Fig.~\ref{fig:appendix_features} visualizes the encode-abstract-decode pipeline. Early encoder layers preserve spatial structure from the concatenated input (both directional blur patterns visible). Through downsampling and deeper layers, features become increasingly abstract. The decoder progressively recovers high-frequency structural detail. Channel-standard-deviation maps (``activity maps'') show consistently high diversity at edge and texture regions, confirming that the network allocates processing effort where structural content is richest.

%% ============================================================
\bibliographystyle{elsarticle-num}
\bibliography{references}

\end{document}